\pdfoutput=1
\documentclass[sigconf]{acmart}

\AtBeginDocument{%
  \providecommand\BibTeX{{%
    \normalfont B\kern-0.5em{\scshape i\kern-0.25em b}\kern-0.8em\TeX}}}

\copyrightyear{2022} 
\acmYear{2022} 
\setcopyright{rightsretained} 

\acmConference[MM '22]{Proceedings of the 30th ACM International Conference on Multimedia}{October 10--14, 2022}{Lisboa, Portugal}
\acmBooktitle{Proceedings of the 30th ACM International Conference on Multimedia (MM '22), Oct. 10--14, 2022, Lisboa, Portugal}
\acmISBN{978-1-4503-9203-7/22/10}
\acmDOI{10.1145/3503161.3548311}

\acmPrice{15.00}
\usepackage{etoolbox}
\makeatletter
\patchcmd{\maketitle}{\@copyrightpermission}{
   \begin{minipage}{0.3\columnwidth}
     \href{https://creativecommons.org/licenses/by/4.0/}{\includegraphics[width=0.90\textwidth]{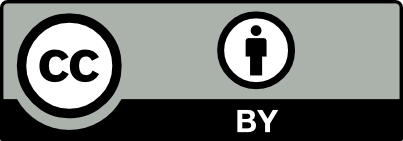}}
   \end{minipage}\hfill
   \begin{minipage}{0.7\columnwidth}
     \href{https://creativecommons.org/licenses/by/4.0/}{This work is licensed under a Creative Commons Attribution International 4.0 License.}
   \end{minipage}
  
   \vspace{5pt}
}{}{}

\makeatother

\usepackage{graphicx}
\usepackage{tikz}
\usepackage{comment}
\usepackage{amsmath} % define this before the line numbering. ,amssymb
\usepackage{color}
\usepackage{graphicx}
\usepackage{bm}
\usepackage{multirow}
\usepackage{makecell}
\usepackage{bbding}
\usepackage{algorithm}
\usepackage{algorithmicx}
\usepackage{algpseudocode}
\usepackage{multirow}
\usepackage{booktabs}
\usepackage{tablefootnote}
\usepackage{threeparttable}
\usepackage{pifont}
\newcommand{\cmark}{\ding{51}}

\acmSubmissionID{2453}

\begin{document}

\title{Generalized Inter-class Loss for Gait Recognition}

%\author{
%          \IEEEauthorblockN{Weichen Yu$^{1,2}$, Hongyuan Yu$^{1}$, Yan Huang$^1$, Liang Wang$^{1*}$} \\
%          \IEEEauthorblockA{$^1$ Center for Research on Intelligent Perception and Computing, \\ Institute of Automation, Chinese Academy of Sciences} \\
%          \IEEEauthorblockA{$^2$ School of Artificial Intelligence, University of Chinese Academy of Sciences} \\
%          \IEEEauthorblockA{\href{mailto:yuweichen16@mails.ucas.ac.cn}{yuweichen16@mails.ucas.ac.cn}, \href{hongyuan.yu@cripac.ia.ac.cn}{hongyuan.yu@cripac.ia.ac.cn}, \href{mailto:yhuang@nlpr.ia.ac.cn}{yhuang@nlpr.ia.ac.cn}, \href{mailto:wangliang@nlpr.ia.ac.cn}{wangliang@nlpr.ia.ac.cn}
%          }
%      }
%\thanks{* Corresponding author}
%\thanks{- This work is partially done in Watrix}
\author{Weichen Yu}
 \authornote{This work was partly done in Watrix.}
 \email{yuweichen16@mails.ucas.ac.cn}
 \orcid{0000-0003-0224-0182}
 \affiliation{%
   \institution{Center for Research on Intelligent Perception and Computing, Institute of Automation, Chinese Academy of Sciences}
   %\streetaddress{P.O. Box 1212}
   \city{Beijing}
%   %\state{Ohio}
   \country{China}
%   %\postcode{43017-6221}
 }
 \author{Hongyuan Yu}
 \email{hongyuan.yu@cripac.ia.ac.cn}
 \orcid{0000-0003-4208-1200}
 \affiliation{%
   \institution{Center for Research on Intelligent Perception and Computing, Institute of Automation, Chinese Academy of Sciences}
%   %\streetaddress{P.O. Box 1212}
   \city{Beijing}
%   %\state{Ohio}
   \country{China}
%   %\postcode{43017-6221}
 }
 \author{Yan Huang}
 \email{yhuang@nlpr.ia.ac.cn}
 \orcid{0000-0002-8239-7229}
 \affiliation{%
   \institution{Center for Research on Intelligent Perception and Computing, Institute of Automation, Chinese Academy of Sciences}
%   %\streetaddress{P.O. Box 1212}
   \city{Beijing}
%   %\state{Ohio}
  \country{China}
%   %\postcode{43017-6221}
 }
 \author{Liang Wang}
 \email{wangliang@nlpr.ia.ac.cn}
 \authornote{Corresponding author.}
 \orcid{0000-0001-5224-8647}
 \affiliation{%
   \institution{Center for Research on Intelligent Perception and Computing, Institute of Automation, Chinese Academy of Sciences}
%   %\streetaddress{P.O. Box 1212}
   \city{Beijing}
%   %\state{Ohio}
   \country{China}
%   %\postcode{43017-6221}
 }
%\settopmatter{printacmref=false} %remove ACM reference format
%\renewcommand\footnotetextcopyrightpermission[1]{}%remove ACM copyright...
%\renewcommand{\shortauthors}{W Yu, H Yu, Y Huang and L Wang}
\renewcommand{\shortauthors}{Weichen Yu, Hongyuan Yu, Yan Huang, \& Liang Wang}
\begin{abstract}
Gait recognition is a unique biometric technique that can be performed at a long distance non-cooperatively and has broad applications in public safety and intelligent traffic systems. Previous gait works focus more on minimizing the intra-class variance %to address cross-view and cross walking condition problems
while ignoring the significance in constraining inter-class variance. To this end, we propose a generalized inter-class loss which resolves the inter-class variance from both sample-level feature distribution and class-level feature distribution. Instead of equal penalty strength on pair scores, the proposed loss optimizes sample-level inter-class feature distribution by dynamically adjusting the pairwise weight. Further, in class-level distribution, generalized inter-class loss adds a constraint on the uniformity of inter-class feature distribution, which forces the feature representations to approximate a hypersphere and keep maximal inter-class variance. In addition, the proposed method automatically adjusts the margin between classes which enables the inter-class feature distribution to be more flexible.
The proposed method can be generalized to different gait recognition networks and achieves significant improvements. We conduct a series of experiments on CASIA-B and OUMVLP, and the experimental results show that the proposed loss can significantly improve the performance and achieves the state-of-the-art performances. %Especially in the cross cloth (CL) condition, the proposed method achieves an accuracy of 89.8\% on CASIA-B (CL).
\end{abstract}




\begin{sloppypar}
\maketitle
\end{sloppypar}
%\vspace{-2em}
\section{Introduction}
Gait recognition is a biometric technique based on unique walking patterns of pedestrians. Compared to other biometrics such as face, fingerprint or iris, gait can be captured at a distance without the cooperation of subjects or intrusion to them. Therefore, it has been applied to many applications recently, such as crime prevention, forensic identification, public security, and intelligent traffic systems \cite{wan2018survey,rida2019robust,deligianni2019emotions}. 

Compared to other image recognition tasks, gait recognition has the challenge of larger intra-class variance and smaller inter-class variance. 
%The large intra-class variance has been widely studied, and many effective works have been proposed to address the large visual difference in intra-class cross-view problem \cite{wu2016comprehensive,zhang2021cross,xu2020cross} and intra-class cross walking conditions problem \cite{zhang2019gait,lin2021gait,fan2020gaitpart}. However, the small inter-class variance is of equal importance in realistic gait recognition. 
The large intra-class variance is mainly due to the large visual dissimilarity among silhouettes from different angles and the same pedestrian wearing different clothes. The small inter-class variance is mainly caused by the following reason: although different pedestrians have different walking patterns, their gait silhouette frames are very similar because pedestrians have the same body structures (head, torso, arms and legs), similar body proportions, and similar walking postures including raising one leg, stepping forward, shifting weight from one leg to another, etc. Many effective works addressing intra-class variance problem have been proposed, such as cross-view gait recognition \cite{wu2016comprehensive,zhang2021cross,xu2020cross} and cross-cloth gait recognition \cite{zhang2019gait,lin2021gait,fan2020gaitpart}, and achieve promising performances. However, most previous works overlook the importance of small inter-class variance and only address it implicitly.

Prior gait recognition works \cite{han2005individual,shiraga2016geinet,chao2021gaitset,hou2020gait,wolf2016multi,lin2021gait} mainly focus on resolving intra-class variance by designing network architectures, while incidentally addressing the small inter-class variance by extracting fine-grained features from the designed networks. 
%overlook the problem of , which 
%indicates the inter-class feature distribution is not carefully constrained, and identity-relevant information can be better preserved.
%But different from intra-class variance, researchers have paid less attention to the problem of inter-class variance.
%To address this problem, 
Previous approaches in gait recogition are categorized into two types: model-based \cite{an2020performance,li2020jointsgait,liao2020model} and appearance-based \cite{han2005individual,shiraga2016geinet,chao2021gaitset,hou2020gait,wolf2016multi,lin2021gait}. The model-based methods use three-dimensional models and convey more information than two-dimensional ones, thus magnifying the inter-class variance. But model-based methods are highly dependent on pose estimation accuracy. And appearance-based methods including Gait Energy Image (GEI)-based, set-based, and 3DCNN-based, extract fine-grained features which enlarge inter-class variance. But the above approaches do not explicitly constrain inter-class feature distribution. 
\begin{figure}[t]
	\centering
	%\vspace{0.3cm}
	\includegraphics[width=0.5\textwidth]{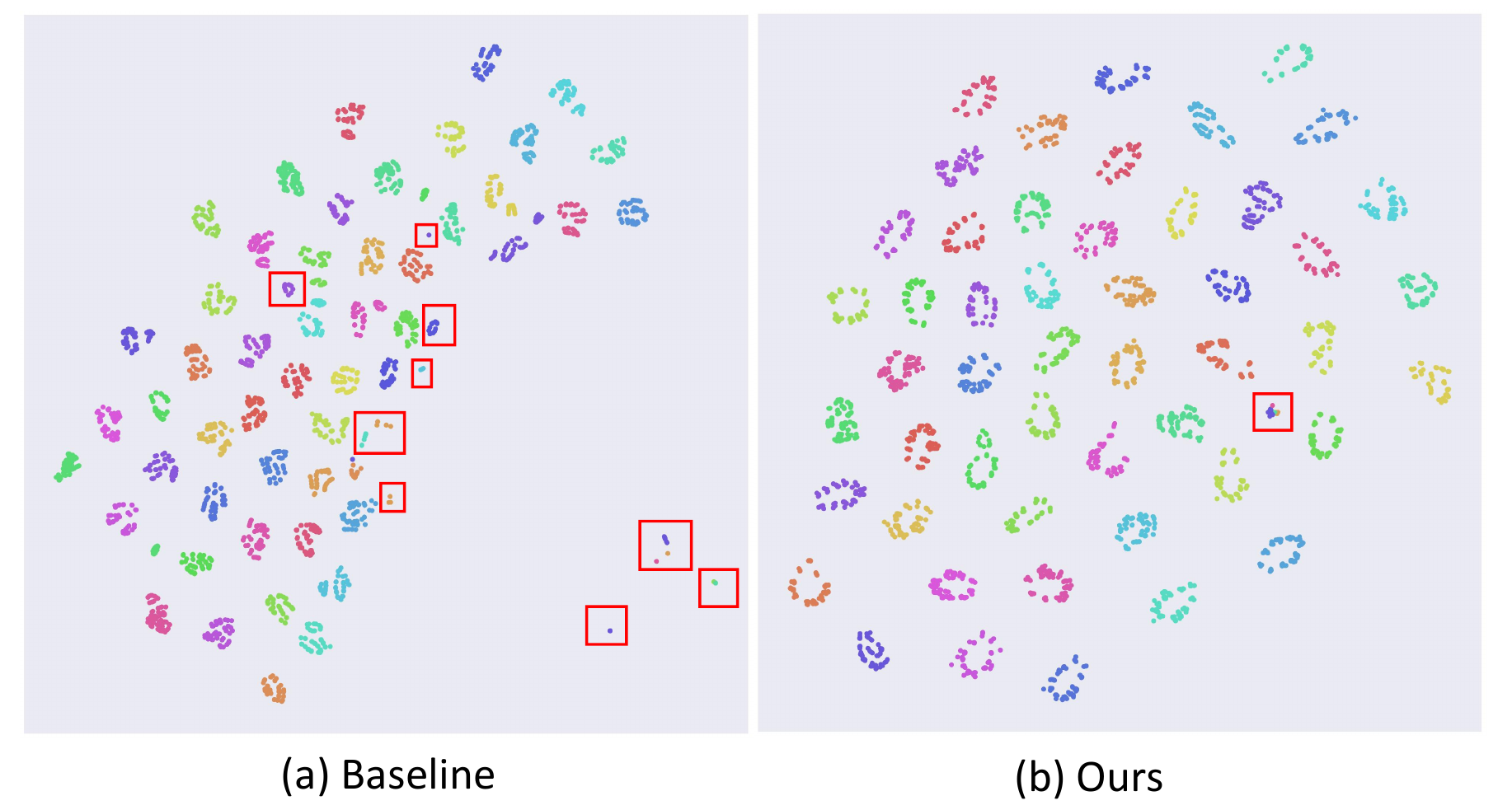}
	\caption{TSNE visualization on CASIA-B test set. One color denotes a class. The red boxes denote suboptimal representations. (a) baseline method. (b) the proposed generalized intra-class loss.}
	\label{fig:vis}
	\vspace{-1em} 
\end{figure}

To address the small inter-class variance, explicitly constraining the inter-class feature distribution is beneficial. From sample-level perspective, some samples of the same viewpoint from different classes are close to each other due to visual similarity. They need to be emphasized to increase their distances and thus the inter-class variance increases. However, previous gait works \cite{chao2019gaitset,chao2021gaitset,fan2020gaitpart,huang20213d,lin2021gait,Li_2020_ACCV} usually treat pairs from different classes inflexibly, where the penalty strength on pair scores is restricted to be equal. 
From class-level perspective, constraining the inter-class feature distribution to be more uniform can increase inter-class variance. Previous works \cite{li2020gait,li2020gait_PR,xu2020cross,lin2020gait,9413894,lin2021multi,li2022spatio} seldom have constraints on inter-class distribution, resulting in lack of spatial symmetry, which is not optimal in keeping maximal mutual information. Nevertheless, margin aims to constrain the distance between classes, but prior works treat all pedestrians equally with the same given margin \cite{chao2019gaitset,chao2021gaitset,fan2020gaitpart,huang20213d}, which lacks flexibility for optimization. Also, different classes with the same given margin lack ability to discriminate between each other.

To this end, we propose a generalized inter-class loss to resolve the inter-class variance problem from both sample-level and class-level. From sample-level perspective, the proposed generalized inter-class loss treats different pairs with dynamic and automatic coefficients, which enables different inter-class samples to dynamically adjust their distances from the anchor class. 

Further, from class-level perspective, the proposed generalized inter-class loss adds a constraint on uniformity of inter-class feature representation and has advantages threefold. Firstly, uniformity prefers the inter-class feature distribution that preserves maximal information. The proposed similarity cross entropy (SimCE) in generalized inter-class loss can be regarded as a variation of von Mises-Fisher kernel density estimation \cite{gopal2014mises,hasnat2017mises,wang2020understanding}, and forces the inter-class feature distribution to approximate a hypersphere in high dimension space. Thus, inter-class uniformity enables maximal inter-class variance. Secondly, the proposed loss is robust with respect to inter-class feature representation differences in its local area. Thirdly, to address the fixed given margin between different classes, the proposed generalized inter-class loss enables automatically adjusting margins between different classes and forces a flexible inter-class feature distribution.

Fig.\ref{fig:vis} is the TSNE visualization of test features. It can be clearly seen that the feature distribution is more uniform, those hard exemplars are effectively optimized and the suboptimal representations in red boxes are decreasing. %Also, every class cluster in (b) is bigger than in (a), which increases the intra-class flexibility and alleviates overfitting.

The contributions of the proposed method are summarized as follows:
\begin{itemize}
% \item{\textbf{Automatic sample-level }}: The proposed reweighting triplet loss helps heuristic dynamic mining of hard samples automatically.
% \item{\textbf{Generalization}}: The proposed SimCE is able to alleviate noise overfitting, improve the uniformity of the induced distribution of the features in high dimension Euclidean space, and improves the optimization direction.
\item{} We propose a unified method to resolve the inter-class variance of gait features from both sample-level and class-level, which dynamically and automatically adjusts the penalty strength on pair scores and margins between different classes.
\item{} We further analyze the properties of the proposed method from three aspects, namely inter-class hard mining, uniformity and robustness of inter-class feature distribution, and dynamic margin. And we illustrate how these properties constrain a better inter-class feature distribution.
\item{} The proposed gait recognition method improves the performance regardless of model structure. Experimental results on public datasets CASIA-B and OUMVLP achieve state-of-the-art performances, especially with an improvement (6.2\%) in different cloth (CL) condition.
\end{itemize}

\section{Related Works}
\subsection{Gait Recognition}
Gait recognition \cite{sarkar2005humanid,hou2020gait,chao2021gaitset,wu2016comprehensive,pan2020optimization,Li_2020_ACCV} is to learn the unique spatio-temporal pattern about the human gait characteristics to obtain its identity information. The gait model input is bipartite: 3D based methods \cite{ariyanto2011model,bodor2009view,zhao20063d,an2020performance} reconstructing the human 3D models from different cameras views, while 2D gait data \cite{li2020gait,li2020gait_PR,song2019gaitnet} is more convenient and easier to achieve. In early gait recognition, to deal with the large variance in gait representation of same identity, hand-crafted view-invariant feature \cite{goffredo2009self,liu2011joint,jean2009towards} and View Transformation Model (VTM) \cite{xing2016complete,kusakunniran2010support} are proposed. 
Recent deep gait recognition networks in CNN are mostly used to capture gait information. GEInet \cite{shiraga2016geinet} and siamese gait network \cite{zhang2016siamese} work on GEI input with CNN. Temporal information capturing includes compressing the gait sequence into one frame using order-consistent statistic operations along temporal dimensions \cite{chao2019gaitset,hou2020gait}. Temporal information is also captured by LSTM or GRU to aggregate pose features in time series to generate the final gait feature \cite{zhang2019gait}. 

To further improve the spatial temporal gait representation, Zhang et al. \cite{zhang2019cross} utilizes a temporal attention mechanism and adaptively adjusts the weights of different frames. GaitNet \cite{zhang2020learning,zhang2019gait} and ICDNet \cite{li2020gait} emphasize disentangled representation learning. GAN is also utilized \cite{chen2021multi,yu2017gaitgan} to generate more data and help with feature constructing. SelfGait \cite{9413894} uses self-supervised learning to perform gait recognition. Gait in the wild attracts researchers' attention \cite{zhu2021gait,zhang2022realgait}, which focuses on real gait conditions and provides datasets in the wild.
However, most of the works above focus more on addressing large intra-class variance and seldom consider small inter-class variance, which is of the same importance as well. 

\subsection{Metric Learning in Gait recognition}
In gait recognition, the most popular loss includes triplet loss, CE classification loss, contrastive loss and hybrid loss. CE classification loss in gait recognition \cite{wu2016comprehensive,hou2020gait,sepas2020view,an2020performance,yu2017gaitgan,wu2015learning,takemura2017input,9616392_Dong,What_Transferred_Dong_CVPR2020} is mainly utilized after prediction head, and takes low-dimension classification logits as inputs to predict classification outputs. In recent years, triplet loss gains popularity in state-of-the-art gait recognition methods \cite{chao2019gaitset,chao2021gaitset,fan2020gaitpart,huang20213d,lin2021gait,Li_2020_ACCV,li2020gait,li2020gait_PR,xu2020cross,lin2020gait,9413894,lin2021multi,li2022spatio}, which compares a baseline input (anchor) to a positive sample with the same identity, and a negative sample with a different identity. Triplet loss function aims to reduce the dissimilarity between feature vectors from the same subject and increase dissimilarity between feature vectors from different subjects. Gait recognition contrastive loss \cite{li2019joint,li2020gait,zhang2016siamese} is usually in the form of Fisher discriminant contrastive \cite{ghojogh2020fisher,chen2022improving}. If the samples are anchor and neighbor, they are pulled towards each other. Otherwise, their distance is increased. In other words, this contrastive loss performs like the triplet loss sequentially rather than simultaneously. Hybrid loss is often utilized as a combination of the loss discussed above. Other metrics adjust their loss functions corresponding to the network special modules. GaitNet \cite{song2019gaitnet} has a segmentation network and a recognition network, and designs the corresponding segmentation loss which is a sigmoid of square error of predicted segmentation mask and the ground truth, and a recognition loss which contains an id loss and a siamese loss. ICDNet \cite{li2020gait} metric contains a triplet loss, a contrastive loss and a reconstruction loss which aims to reconstruct the disentangled part. In this paper, we focus more on analyzing the loss properties on inter-class variance in gait recognition task.

\section{Inter-class Feature Distribution}
\subsection{Problem Formulation}
We formulate the inter-class feature distribution loss problem in gait recognition as follows: given a objective function $\mathcal{L}$, and a fixed network architecture, after training to convergence, the loss stays stable, i.e. $\Delta\mathcal{L} \le \epsilon$, we denote the converged network parameters as $W(\mathcal{L})$. Then for every anchor feature $a$, we denote the positive sampled feature as $p$, and negative sampled feature as $n$, where the corresponding $a,p,n$ are computed by input silhouette sequences and $W(\mathcal{L})$. Since we focus on inter-class distribution, relative feature differences $u=a-p$, $v=a-n$ are more informative than a single feature. Thus we do a coordinate conversion. Then to find a inter-class feature distribution is to find a solution in Eq.\ref{eq:loss_fomulate}.
\begin{equation} \label{eq:loss_fomulate}
{\min_{\mathcal{L}} \mathcal{L}_{eval}(a,u,v,W(\mathcal{L}))} 
\end{equation}
Usually the loss $\mathcal{L}$ on evaluation/test dataset is the same as on train dataset $\mathcal{L}$, and in the following context we do not differentiate the two of them. An ideal objective $\mathcal{L}$ generalizes well and is small on both train set and test set. Note that it is small but not zero because the dataset inevitably contains noise. 

In this paper, we explore how a loss interacts with inter-class feature distribution in gait recognition task, and we carefully analyze the properties of objectives in gait recognition from three aspects: explicit inter-class hard mining, uniformity and robustness of inter-class distribution, and dynamic margin.

%and cross-entropy classification loss \cite{wang2022gaitstrip,li2022strong,huang20213d,Li_2020_ACCV,lin2021gait,lin2021multi,zhang2019cross} together dominate the field and their effectiveness is verified by multiple works. 
%In addition, mining hard relations is a significant and difficult topic for loss design, and triplet loss as well as cross-entropy classification loss already have respective way to address hard-mining problem in different manners. 
% However, triplet loss and cross-entropy classification loss have different definition equations and thus different intrinsic properties and respective limitations, and can be better designed to address gait recognition task. 

\subsection{Explicit Inter-class Hard Mining}
According to Eq.\ref{eq:loss_fomulate}, we firstly analyze the \textbf{first order gradient} of $\mathcal{L}$ on $v$, which actually indicates hard relations mining. The first order gradient on $v$ affects the sample-wise inter-class feature distribution: gradient on $v$ is to push it with penalizing strength to be further from anchor point, and pushing the smaller $v$ stretches the hard classes to discriminate from each other and increases inter-class variance. %To attain discriminative representations between different classes, it is effective to push the anchor-negative pairs from harder class with larger penalty strength. 
Recent gait recognition works utilizes vanilla triplet loss \cite{chao2019gaitset,chao2021gaitset,fan2020gaitpart,huang20213d,lin2021gait,Li_2020_ACCV,li2020gait,li2020gait_PR,xu2020cross,lin2020gait,9413894,lin2021multi,li2022spatio} as a prevailing objective (Eq.\ref{eq:triloss}), which equally treats all triplets and ignores the importance of mining hard relations. FaceNet \cite{schroff2015facenet} addresses this by selecting the hardest negative exemplars within a batch. However, this solution brings two doubts. Firstly, as Hermans et al. \cite{hermans2017defense} state, selecting all positives is more stable compared to selecting the hardest positives. %since data noise is prevalent in gait dataset because of preprocessing, the hardest exemplars have a large possibility to be noisy, and using the hardest only samples leads to severe noise overfitting. 
%Samples of different classes have different semantic similarities, and  the differences between anchor class and two negative samples from other classes. During neural network training, hard samples contain valuable information for feature representation and classification. In 
%and in vanilla triplet loss, the hard mining lies in the hard anchor-positive and anchor-negative relation. The  triplet loss as in  equally treats all triplets and ignores the importance of mining hard relations. FaceNet \cite{schroff2015facenet} addresses this by selecting the hardest negative exemplars within a batch. However, this solution brings two doubts. Firstly, since data noise is prevalent in gait recognition, the hardest exemplars have a large possibility to be noisy, and using the hardest only leads to severe noise overfitting. 
And we analyze the possible reason is the lack of triplets selected: hardest-only principle significantly reduces the number of triplets and increases the possibility to overfit out-of-distribution exemplars, which are prevalent in gait dataset because of preprocessing. Secondly, to obtain a meaningful representation of the distances, FaceNet \cite{schroff2015facenet} samples numerous faces per identity \footnote{In FaceNet \cite{schroff2015facenet}, batch size is 1800.}, which is impractical in gait recognition because of the dataset size\footnote{In gait recognition, batch size is usually 64 or 128.}. Despite the large batch size, selecting the hardest negatives can in practice lead to bad local minima early in training, specifically, it results in a collapsed model. Thus, instead of selecting the hardest exemplars, it is straightforward to weight each anchor-positive and anchor-negative pair, i.e., to down-weight the loss assigned to well-classified pairs.
\begin{equation} \label{eq:triloss}
{\mathcal{L}_{tri}} = ReLU(m+dist(u)-dist(v))
\end{equation}
where $m$ is the margin. In gait recognition task, $dist(\cdot)$ is usually Euclidean distance $||\cdot||_2$.
% \begin{figure}[t]
% 	\centering
% 	\includegraphics[width=0.5\textwidth,height=5cm]{fig/order.pdf}
% 	\vspace{-2em}
% 	\caption{An example of noise overfitting. Different colors represent different classes. The circles are normal samples, triangles are samples with some visual dissimilarities, and the stars are noisy samples. The black samples and the arrows indicate the moving directions after training for some epochs. (a) feature distribution after the early training period. (b) feature distribution after triplet loss training, and overfitting occurs. (c) the proposed loss helps alleviate noise overfitting.}
% 	\label{fig:margin}
% 	\vspace{-2em}
% \end{figure}

To assign different weights automatically to different triplets, we adopt similarity matrix $\mathcal{S}$ as in Eq.\ref{eq:s-triloss} for the following reasons. Firstly, inspired by focal loss \cite{lin2017focal} which utilizes a function of the input of the loss to re-weight every loss term, the similarity matrix can also be computed online from the input of triplet loss and requires no additional modules or alternating the network structure, which is simple and can be easily generalized. Secondly, similarity matrix, although simple in its form, is a good reflection of pair distance, and the normalized form of similarity matrix $\mathcal{S}/\mathcal{S}_{max}$ is informative for pair comparison in a batch. We mine hard relations from a granularity of pairs, instead of triplets or batches, which is more flexible and detailed.
%note that the contribution of easy triplets to loss are non-trivial when summed up in a batch
\begin{equation} 
\begin{split}
\label{eq:s-triloss}
    {\mathcal{L}_{s-tri}} &= ReLU(m+w_{a,p}dist(a,p)-w_{a,n}dist(a,n)) \\
&=ReLU(m+f(\mathcal{S}_{a,p})dist(a,p)-f(\mathcal{S}_{a,n})dist(a,n)) \\
\end{split}
\end{equation}
where the $w_{a,p}$, $w_{a,n}$ are the re-weighting terms, and $\mathcal{S}_{a,p}$, $\mathcal{S}_{a,n}$ are the corresponding entries in the similarity matrix, and $f(\cdot)$ here is a function inversely proportional to $\mathcal{S}$. We adopt a linear $f(x) = (1-x)/2$. More carefully designed form of $f(\cdot)$ is sure to further improve the performance, but it is not the priority of this paper.

Since $f(\cdot)$ is a function inversely proportional to $\mathcal{S}$, it can automatically mine hard triplet pairs. If the anchor-positive pair are well-classified, $\mathcal{S}_{a,p}$ is rather big, $w_{a,p}$ is thus small, and this term in final triplet loss is rather small, and vice versa. If the anchor-negative pair are well-classified, $\mathcal{S}_{a,n}$ is rather small, and $w_{a,n}$ is thus big, and this term in final triplet loss is rather small, vice versa. In this way, without changing network structures and the computation cost is rather small, we mine hard triplets automatically by assigning lower weights to well-classified triplets.

To notify, triplet loss and CE classification loss implicitly mines hard examples because the gradient of triplet loss and CE classification loss are proportional to the difference between pairs. The implicit hard mining is an intrinsic property of loss and most losses have this implicit property but lack careful loss design.

%Such an optimization manner is inflexible, because the penalty strength on every single similarity score
%is restricted to be equal. Our intuition is that if a similarity score deviates far from the optimum, it should be emphasized.
\begin{figure}[t]
	\centering
	%\vspace{0.3cm}
	\includegraphics[width=0.5\textwidth,height=4cm]{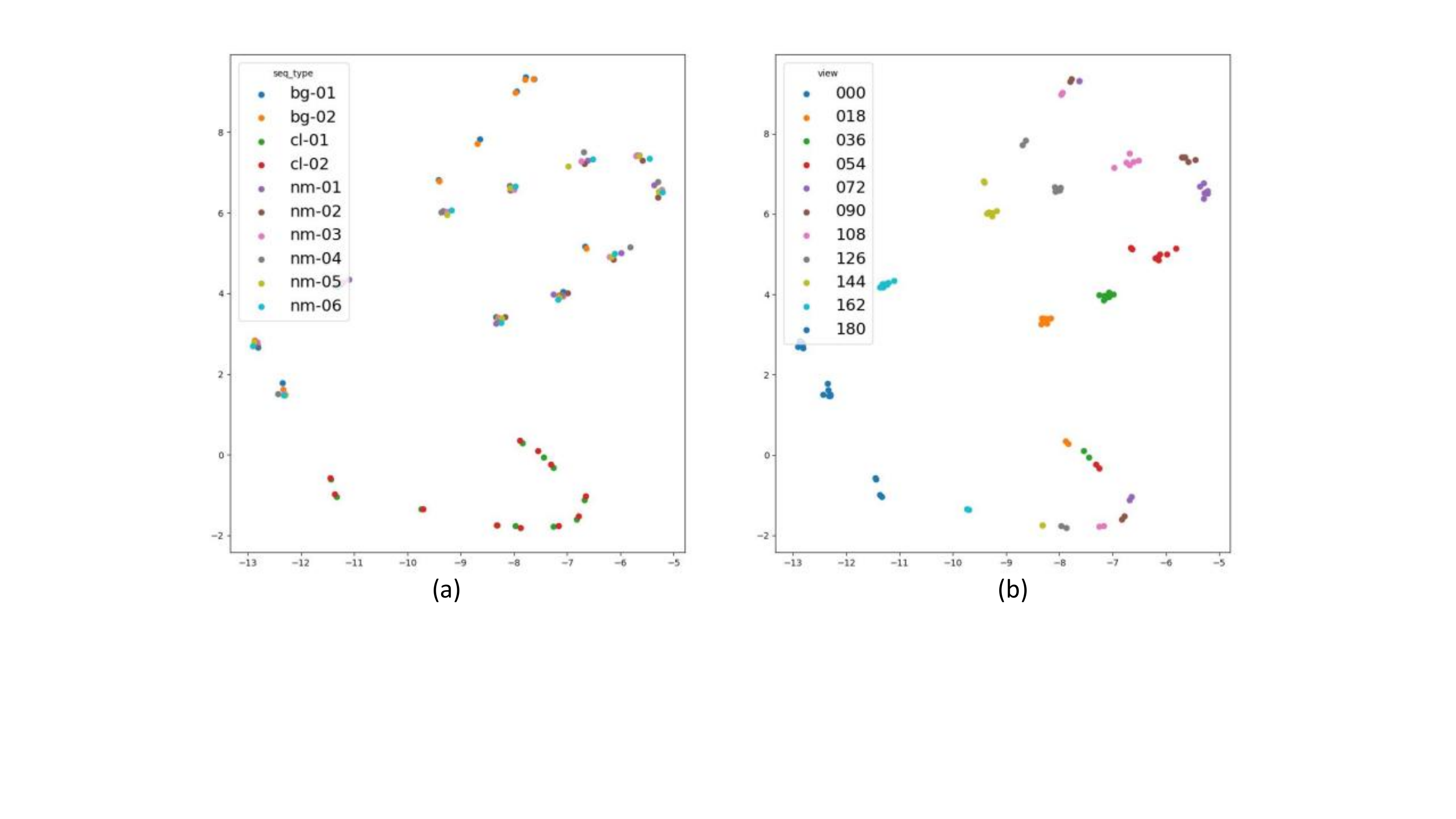}
	\vspace{-2em}
	\caption{Visualization of a exemplar identity `080' in CASIA-B test set. Every point denotes a sequence feature of `080'. (a) different color denotes different walking conditions. And CL sequences separate from NM and BG sequences. (b) different color denotes different views. And sequences of the same views are close to each other.}
	\label{fig:intraclass}
	\vspace{-2em} 
\end{figure}

\subsection{Uniformity and Robustness of Inter-class Distribution}
\subsubsection{Uniformity and Hypersphere}
Improving uniformity of inter-class feature distribution is beneficial for increasing inter-class variance. Uniformity prefers a feature distribution that preserves maximal information. In addition, Wang et al. \cite{wang2020understanding} points out that well-clustered spherical spaces are linearly separable, while this does not hold for Euclidean spaces. Thus we expect the feature distribution to approach a hypersphere and increase separability. Prior gait recognition lacks constraints on inter-class features to be uniformly distributed in the feature space (Fig.\ref{fig:vis}(a)).
%In gait recognition, objective utilizing Euclidean distance as $dist(\cdot)$ is considered unquestionable and popular in recent deep learning gait recognition work \cite{chao2019gaitset,chao2021gaitset,fan2020gaitpart,huang20213d,lin2021gait,Li_2020_ACCV,li2020gait,li2020gait_PR,xu2020cross,lin2020gait,9413894,lin2021multi,li2022spatio}. However, Wang et al. \cite{wang2020understanding} points out that well-clustered spherical spaces are linearly separable, while this does not hold for Euclidean spaces. More importantly, uniformity prefers a feature distribution that preserves maximal information, but inter-class feature representations learned by existing methods are not constrained to be uniformly distributed in the feature space (Fig.\ref{fig:vis}(a)). And in the feature space, lack of uniformity reflects small differences between classes and non-discriminative feature representations learned by models. And the linearly inseparable property impedes the classifier to predict with high accuracy.

%Thus, we propose to add a term utilizing similarity in cosine form to constrain the feature representation to be more uniform and approach a hypersphere in the high dimension gait space. 
Previous works utilizes similarity in cosine form to constrain the feature representation to be more uniform, such as congenerous cosine loss (CoCo Loss) \cite{liu2017rethinking} introduces cosine loss to center loss in image classification on CIFAR, and Large Margin Cosine Loss (LMCL) \cite{wang2018cosface} introduces a cosine and a margin to classification loss in face recognition. However, when combined with gait recognition, CoCo Loss and LMCL utilize class center as the "ground truth" and force every feature to be close to the center, which is inferior for the following reason. As shown in Fig.\ref{fig:intraclass}, in every gait class of one identity, there is an intra-class structure since the sequences of the same view and the same walking conditions are visually similar and stay close to each other. And the network inevitably keeps features extracted from front convolution layers and preserves visual similarity information. Simply approximating every feature by center feature results in destruction of the intra-class structure. Since the id-relevant information is not totally disentangled with id-irrelevant information (e.g., views, walking conditions), destruction of the intra-class structure throws features extracted from front convolution layers away as well as ignoring the valuable information in these features. Thus, we propose to keep the intra-class structure without using class centers, and propose the cross entropy of similarity matrix between features representations (SimCE loss) as in Eq.\ref{eq:3-nceloss}. We will carefully analyze why this SimCE loss helps with inter-class variance. 
\begin{equation} \label{eq:3-nceloss}
{\mathcal{L}_{SimCE}(a,u,v)} = -log {\frac{e^{a(a-u)/T}}{e^{a(a-u)/T}+{e^{a(a-v)/T}}}} %= log(1+\frac{e^{a(/T}}{e^{ap/T}}})
\end{equation}
\\\textbf{Hypersphere Embedding.} A mixture of von Mises-Fisher kernel density estimations (Eq.\ref{eq:vmf}) constructs a hypersphere embedding. And the form ${\frac{e^{ap/T}}{e^{ap/T}+{e^{an/T}}}}$ can be explained by a von Mises-Fisher Mixture Model (vMFMM) \cite{gopal2014mises,hasnat2017mises,wang2020understanding}. Training with ${\mathcal{L}_{SimCE}}$ minimizes every $\kappa$ and $\kappa$ in vMF function determine uniformity of random variables distribution on a sphere. The smaller the $\kappa$ is, the more uniform the distribution is. Note that we generalize the $\mu$ from $||\mu||_2 = 1$ to $\mu$ in Euclidean space. Thus, beyond the triplet loss which utilizes Euclidean constraints in high dimension space, adding SimCE loss provides an additional constraint that forces the features to be more similar to uniform hypersphere embeddings. 
\begin{equation} \label{eq:vmf}
{\mathcal{V}_{d}(x|\mu,\kappa)} = C_d(\kappa)exp(\kappa\mu^Tx) 
\end{equation}
where, $\mu$ denotes the mean and $\kappa$ denotes the concentration parameter (with $\kappa \geq 0$). $C_d(\kappa) = \frac{\kappa^{d/2-1}}{(2\pi)^{d/2}I_{d/2-1}(\kappa)}$ is the normalization constant, where $I(\cdot)$ is the modified Bessel function of the first kind. The shape of the vMF distribution depends on the value of the concentration parameter $\kappa$. 

Euclidean triplet loss forces the features to have radius disparity and is radius discriminative, while SimCE loss forces the features to have tangential disparity, and is tangential separable. Thus these two losses together results in a more uniformly distributed feature space, as in Fig.\ref{fig:vis}(b).% , adding this additional SimCE loss helps class clusters to better occupy the whole space.

% \begin{table*}[t]
% \caption{Loss analysis}
% \begin{tabular}{|l|l|l|l|l|l|}
% \hline
%                       & Triplet loss & S-Triplet loss  & CE classification loss & SimCE loss  & Contrastive loss                                                                                               \\ \hline
% Range of optimization & 2 classes, 3 samples & 2 classes, 3 samples & All C classes, all samples & 2 classes, 3 samples & \begin{tabular}[c]{@{}l@{}}Several classes, \\ Several samples\end{tabular} \\ \hline
% Mining hard relations & Implicit             & Explicit             & Implicit                   & Implicit  & Implicit                                                                                               \\ \hline
% Constraint                & Euclidean            & Cosine and Euclidean & Cosine                     & Cosine & Cosine                                                                                                  \\ \hline
% \end{tabular} \label{tab:loss}
% \end{table*}
\subsubsection{Robustness on Local Area}
Following the function Eq.\ref{eq:loss_fomulate}, we further analyze the property of \textbf{second order gradient} on $v$ with a fixed $a$, which indicates the robustness of the objective. And the objective function expectation on local area becomes: 
\begin{equation} \label{eq:robust}
\begin{split}
&E_{\delta \sim U_{[-\epsilon,\epsilon]}}\mathcal{L}(a,u,v+\delta) \\
&\approx E_{\delta \sim U_{[-\epsilon,\epsilon]}} [\mathcal{L}(a,u,v)+\delta \nabla_{v}{L}(a,u,v) +1/2\delta^T\nabla_{v}^2{L}(a,u,v)\delta] \\
&=\mathcal{L}(a,u,v)+\epsilon^2/6Tr{\nabla_{v}^2{L}(a,u,v)}
\end{split}
\end{equation}
where the second term in Eq.\ref{eq:robust} is canceled out since $E[\delta] = 0$ and the off-diagonal elements of the third term becomes 0 after taking the expectation on $\delta$. 

If Hessian is stable in a local region of ${L}(a,u,v)$, then the quantity of  Eq.\ref{eq:robust} can approximately bound the performance drop when suffering perturbation of unavoidable data noise or evaluation on test dataset, which improves robustness. As shown in Eq.\ref{eq:nceloss-hessian}, the trace of the Hessian of the proposed $\mathcal{L}_{SimCE}$ is bounded within 1, while in Eq.\ref{eq:triloss-hessian}, the trace of Hessian has no such bound in range of $[0,+\infty)$. The analysis indicates that triplet loss $\mathcal{L}_{tri}(a,u,v)$ has less robustness compared to the proposed $\mathcal{L}_{SimCE}(a,u,v)$.
% \begin{equation} \label{eq:triloss-gradient}
% \begin{split}
% \frac{\partial{\mathcal{L}_{tri}}}{\partial{u_i}} &= \frac{\partial{ReLU(m+||u||_2-||v||_2)}}{\partial{u_i}} \\
% &= \frac{u_i}{||u||_2} \  if \  m+||u||_2-||v||_2 \geq 0
% \end{split}
% \end{equation}
\begin{equation} \label{eq:triloss-hessian}
\begin{split}
trace(H_{ij}(\mathcal{L}_{tri})) 
&= \sum_{i}\frac{||v||_2-(v_i)*cos(\theta)}{||v||_2^2}  \\
&\approx (n-1)\frac{1}{||v||_2} \\
& if \  m+||u||_2-||v||_2 \geq 0
\end{split}
\end{equation}

% \begin{equation} \label{eq:3-nceloss-gradient}
% \begin{split}
% \frac{\partial{\mathcal{L}_{SimCE}}}{\partial{v_i}} &= -\frac{\partial{log {\frac{e^{ap}}{e^{ap}+{e^{an}}}}}}{\partial{v_i}} \\
% &= -\frac{a_ie^{a(a-v)}}{e^{a(a-u)}+e^{a(a-v)}}
% \end{split} 
% \end{equation}
\begin{equation} \label{eq:nceloss-hessian}
\begin{split}
&trace(H_{ij}(\mathcal{L}_{SimCE})) \\
&= \sum_{i}\frac{a_i^2e^{a(a-u)}e^{a(a-u)}}{(e^{a(a-u)}+e^{a(a-v)})^2} \\
&= \frac{e^{a(a-u)}e^{a(a-u)}}{(e^{a(a-u)}+e^{a(a-v)})^2} \leq 1/2
\end{split}
\end{equation}
%where $\mathcal{L}_{SimCE}$ is a simplified version for $\mathcal{L}_{SimCE}(a,u,v)$.
\subsection{Dynamic Margin} 
Margin also depicts a property of the inter-class distribution and aims to constrain the distance between classes.
Popular losses in gait recognition and face recognition (e.g. triplet loss \cite{hermans2017defense}, LMCL \cite{wang2018cosface}) require a given fixed margin, which brings three doubts: semantic interpretability, training efficiency and noise overfitting. %SimCE requires no margin and is able to alleviate the above problems. 
As in Eq.\ref{eq:ncetotri}, the SimCE term can be approximately represented as triplet loss with a dynamic margin $(an-ap)^2/T+2T$, and is able to alleviate the above problems. %where the better the classification corresponds to larger margin and the dynamic margin normalizes every triplets to contribute similarly to loss terms and alleviates the noise overfitting.
\begin{equation} \label{eq:ncetotri}
\begin{split}
{\mathcal{L}_{SimCE}} &= -log {\frac{e^{ap/T}}{e^{ap/T}+{e^{an/T}}}} = log(1+exp((an-ap)/T)) \\ &\approx exp((an-ap)/T) \\
&\approx 1+(an-ap)/T +(an-ap)^2/{2T^2} \\
&= 1-1/{2T}(||a-n||^2-||a-p||^2-(an-ap)^2/T) \\
&\propto - (||a-n||^2-||a-p||^2-(an-ap)^2/T-2T)
\end{split}
\end{equation}

Firstly, pushing all classes with the same given margin is not semantic interpretable. For example, class `001' and class `002' are all young females with similar body shapes and walking habits, while `003' is a child and is dissimilar to `001'. If `001' is the anchor, and the same given margin pushes `002' and `003' with the same distance $m$, which lacks semantic interpretability. SimCE loss computes the similarity of feature pairs, and adaptively adjusts the distances between different classes. The larger the differences between $ap$ and $an$, the larger the dynamic margin $(an-ap)^2/T+2T$ is. Thus, the inter-class feature distribution is further optimized.

Secondly, it is also inefficient for training since `002' and `003' are at the same margin $m$ from `001', and it requires further training to separate `002' and `003'. In addition, if batch-hard triplet loss is utilized, it requires computational cost to generate a hard triplet. If the batch size is $[N,K]$, then following the work of hard triplet loss \cite{hermans2017defense}, it needs to compute $(N-1)*K$ pairs to output the hard triplet pairs. SimCE alleviates this by making use of all triplets and separating them with adaptive discriminative distances.

Finally, when the distance of anchor-positive and anchor-negative triplet satisfies the margin, this triplet is not taken into consideration when computing the final loss. Denote the number of nonzero triplets as $N_{non}$. In gait recognition, we observe that $N_{non}$ swiftly decays to a small integer, as in Fig.\ref{fig:loss}, which indicates overfitting. After the early training period %(Fig.\ref{fig:margin}(a)),
the structure of the class cluster is that the clean feature (circles) are the closest to the class center, and difficult dissimilar features (triangles) are further, while the noisy features (stars) are far from the class center. After the network learns gradually, clean samples' contribution to loss is zero, then the network continues to optimize other samples. As the noisy samples gradually reach a higher proportion, %as in Fig.\ref{fig:margin}(b), 
the discriminative boundaries in feature space gets complicated and overfit noise. Moreover, by using a hard given margin, knowledge learned in the early training period is down-weighted. By adding the proposed loss, all features regardless of their distance from others are taken into consideration. At the same time, SimCE also mines hard pairs explicitly, which indicates that the hard exemplars are given with large loss. The proposed method alleviates the problem of noise making up a large proportion in the late training period. 
% \\\textbf{Mathematical Analysis.} Here we also perform a little bit mathematical analysis to give a hint whether these two losses are relatively stable. 

% As for the first order derivative towards $a-p$, triplet loss is larger than SimCE loss (Eq.\ref{eq:gradient-nce} and ). 

To conclude, the proposed method optimizes inter-class distribution for the following reasons: 1) the proposed method has the property to force hypersphere embeddings in feature space with both Euclidean and angular constraints. 2) the proposed method provides a bounded Hessian in a local region of test accuracy and ensures the robustness of loss metric. 3) SimCE loss helps alleviate the degeneration caused by the fixed margin, improves semantic interpretability, training efficiency and alleviates noise overfitting.

\begin{figure}[t]
	\centering
	\includegraphics[width=0.5\textwidth,height=4cm]{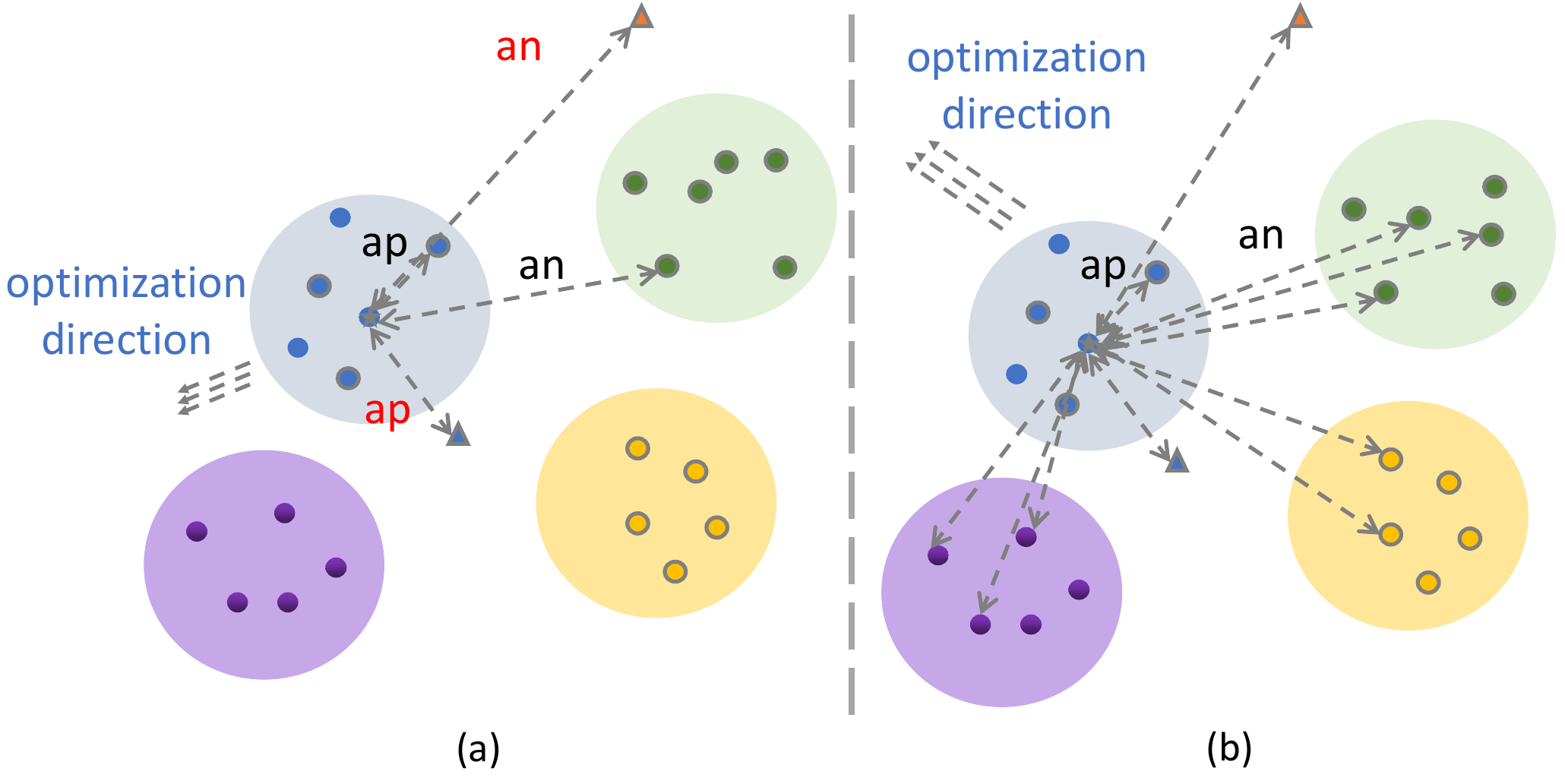}
	\vspace{-2em}
	\caption{An example of optimization. Different colors indicate different classes. Dash lines denote the sampled features for loss calculating. Triangles and font in red denote noisy exemplars and circles and font in black denotes clean ones. (a) SimCE loss. (b) m-SimCE loss.}
	\label{fig:range}
	\vspace{-2em}
\end{figure}
\subsection{From Three Terms to Multiple Terms} \label{sec:optimization range}
Inspired by contrastive loss \cite{khosla2020supervised}, InfoNCE loss \cite{van2018representation} and N-pair loss \cite{sohn2016improved}, which have similar forms to SimCE, we change SimCE with three terms to multiple terms because we aim to utilize multiple negative exemplars for constructing contrastive groups and for feature learning. Thus, m-SimCE every time considers $N*K$ samples of $N$ classes, where $Batch size = [N,K]$.\footnote{$N$ refers to the number of classes, and $K$ refers to the number of sequences sampled from every class.} The advantages and disadvantages of multiple terms are as follows.
\begin{equation} \label{eq:nceloss}
{\mathcal{L}_{m-SimCE}} = -log {\frac{e^{ap/T}}{e^{ap/T}+\sum_{k^-}{e^{an/T}}}}
\end{equation}
\\\textbf{Advantages.} Utilizing SimCE and triplet loss is actually equivalent to a two-class classification. Similar to the work of Zhang et al. \cite{zhang2021beyond}, there is a need to enable the joint optimization of multiple instances within per-query optimization for gait recognition. As illustrated in Fig.\ref{fig:range}, more negative terms bring advantages including robustness to noise and a larger range for optimization. By enlarging the number of samples for every loss term, the influence of noise contributes less to the total loss. And as in Fig.\ref{fig:range}(a), the optimization direction is determined by two terms and is probable to be only optimal for this pair, while in Fig.\ref{fig:range}(b) optimization direction is more optimized.
\\\textbf{Disadvantages.} However, m-SimCE highly depends on intrinsic visual similarity to separate dissimilar features from similar ones. If the intrinsic similarity is large intra-class, the intra-class variance gets larger. In other words, the multiple term loss increases inter-class and intra-class variance simultaneously, and if the intra-class variance of datasets is large, m-SimCE harms the overall classification.
\\\textbf{Guideline.} Considering the advantages and disadvantages above, the guideline is that if the cross cloth condition makes up a high proportion in the dataset, we recommend using ${\mathcal{L}_m}$, and otherwise using ${\mathcal{L}_s}$. Or you can use a hyperparameter to adjust the weight of ${\mathcal{L}_m}$ and ${\mathcal{L}_s}$, which is not the prior of this paper.

Thus, we have two final generalized similarity loss form, combining similarity-triplet loss, cross entropy loss and SimCE loss (or multiple SimCE loss), as in Eq.\ref{eq:loss}
\begin{equation} \label{eq:loss}
\begin{split}
{\mathcal{L}_s} &= \mathcal{L}_{s-tri}+\mathcal{L}_{CE}+\mathcal{L}_{SimCE} \\
{\mathcal{L}_m} &= \mathcal{L}_{s-tri}+\mathcal{L}_{CE}+\mathcal{L}_{m-SimCE}
\end{split}
\end{equation}
\begin{center}
\begin{table*}[!t]
  \scriptsize
  \centering
  \vspace*{-1em}
   \caption{Rank-1 accuracy (\%) on CASIA-B under all view angles, different settings and conditions, excluding identical-view case. \textbf{Bold} and \textit{Italic} fonts indicate the best and second-best results respectively.
  }
  \vspace*{-1em}
  \resizebox{0.98\textwidth}{!}{
    \begin{tabular}{c|c|c|c|c|c|c|c|c|c|c|c|c|c|c}
    \toprule
    \multicolumn{3}{c|}{Gallery NM\#1-4}  &\multicolumn{12}{c}{$0^{\circ}$-$180^{\circ}$} \\
    \hline
    \multicolumn{3}{c|}{Probe}    & $0^{\circ}$     & $18^{\circ}$    & $36^{\circ}$    & $54^{\circ}$    & $72^{\circ}$    & $90^{\circ}$    & $108^{\circ}$   & $126^{\circ}$   & $144^{\circ}$   & $162^{\circ}$   & $180^{\circ}$  & Mean\\
    \midrule

  %  \hline
    \multicolumn{1}{c|}{\multirow{12}[2]{*}{\textbf{ST(24)}}} & \multicolumn{1}{c|}{\multirow{6}[2]{*}{NM\#5-6}} 
%   \multicolumn{1}{c|}{\multirow{12}[2]{*}{\textbf{ST(24)}}} &
    % \multicolumn{1}{c|}{\multirow{9}[18]{*}{\textbf{ST(24)}}} &
    % \multicolumn{1}{c|}{\multirow{5}[10]{*}{NM\#5-6}} 
    & ViDP &  $-$     &  $-$     &   $-$    & 59.1  &   $-$    & 50.2  &   $-$    & 57.5  &    $-$   &   $-$    &  $-$     & $-$ \\
&       & CMCC & 46.3  &   $-$    &   $-$    & 52.4  &    $-$   & 48.3  &     $-$  & 56.9  &    $-$   &    $-$   &      $-$ &  $-$\\
% \cline{3-15}   
&       & CNN-LB & 54.8  &   $-$    &    $-$   & 77.8  &   $-$    & 64.9  &  $-$     & 76.1  &  $-$     &    $-$   &    $-$   & $-$ \\
% \cline{3-15}     
&       & GaitSet & 71.6  & 87.7  & 92.6 & 89.1  & 82.4  & 80.3 & 84.4  & 89.0  & 89.8  & 82.9  & 66.6  & 83.3  \\
% \cline{3-15}   
% &       & Ours  & \RED{72.1} & \RED{83.4} & 88.6  & \RED{88.6} & \RED{80.9} & 74.4  & \RED{82.3} & \RED{88.3} & \RED{89.5} & \RED{86.5} & \RED{69.9} & \RED{82.2} \\
&       & GaitGL (baseline)   & 77.0 & 87.8 & 93.9 & 92.7 & 83.9 & 78.7 & 84.7 & 91.5 & 92.5 & 89.3 & 74.4 & 86.0 \\
&       & \textbf{Ours} (${\mathcal{L}_m}$)  & \textit{80.1} & \textit{90.5} & \textit{95.5} & \textit{93.5} & \textit{84.7} & \textit{80.9} & \textit{86.9} & \textit{91.9} & \textit{94.2} & \textit{90.7} & \textit{76.6} & \textit{87.8} \\
&       & \textbf{Ours} (${\mathcal{L}_s}$)  & \textbf{82.8} & \textbf{92.2} & \textbf{96.2} & \textbf{94.0} & \textbf{86.0} & \textbf{81.3} & \textbf{87.4} & \textbf{93.3} & \textbf{94.9} & \textbf{90.8} & \textbf{77.5} & \textbf{88.7} \\

\cline{2-15}          & \multicolumn{1}{c|}{\multirow{3}[2]{*}{BG\#1-2}} & GaitSet & 64.1  & 76.4  & 81.4  & 82.4  & 77.2  & 71.8  & 75.4  & 80.8  & 81.2  & 75.7  & 59.4  & 75.1  \\
% \cline{3-15} 
% &       & Ours  & \RED{64.2} & \RED{73.8} & \RED{79.3} & \RED{80.8} & \RED{71.3} & \RED{65.3} & \RED{72.3} & \RED{79.2} & \RED{82.5} & \RED{79.7} & \RED{60.3} & \RED{73.5} \\
&       & GaitGL (baseline)   & 68.1 & 81.2 & 87.7 & 84.9 & 76.3 & 70.5 & 76.1 & 84.5 & 87.0 & 83.6 & 65.0 & 78.6 \\
&       & \textbf{Ours} (${\mathcal{L}_m}$)  & \textbf{75.4} & \textit{85.4} & \textit{90.8} & \textit{87.7} & \textit{76.9} & \textit{73.5} & \textbf{79.3} & \textit{86.7} & \textit{89.8} & \textit{86.1} & \textbf{71.3} & \textit{82.1} \\
&       & \textbf{Ours} (${\mathcal{L}_s}$)  & \textit{73.1} & \textbf{87.5} & \textbf{91.3} & \textbf{87.8} & \textbf{77.9} & \textbf{73.6} & \textit{78.6} & \textbf{87.2} & \textbf{90.5} & \textbf{87.2} & \textit{70.9} & \textbf{82.3} \\

\cline{2-15}          & \multicolumn{1}{c|}{\multirow{3}[2]{*}{CL\#1-2}} & GaitSet & 36.4  & 49.7  & 54.6  & 49.7  & 48.7  & 45.2  & 45.5  & 48.2  & 47.2  & 41.4  & 30.6  & 45.2  \\
% \cline{3-15}  
% &       & Ours  & \RED{45.8} & \RED{59.1} & \RED{62.7} & \RED{62.5} & \RED{56.7} & \RED{51.5} & \RED{57.6} & \RED{60.5} & \RED{58.0} & \RED{54.4} & \RED{37.9} & \RED{55.1} \\
    % 
&       & GaitGL (baseline)   & 46.9 & 58.7 & 66.6 & 65.4 & 58.3 & 54.1 & 59.5 & 62.7 & 61.3 & 57.1 & 40.6 & 57.4 \\
&       & \textbf{Ours} (${\mathcal{L}_m}$)  & \textbf{56.8} & \textbf{73.8} & \textbf{79.6} & \textbf{75.3} & \textbf{68.7} & \textit{61.4} & \textbf{68.4} & \textbf{74.0} & \textbf{72.1} & \textbf{65.7} & \textbf{50.4} & \textbf{67.8} \\
&       & \textbf{Ours} (${\mathcal{L}_s}$)  & \textit{52.3} & \textit{68.8} & \textit{75.3} & \textit{72.3} & \textit{68.3} & \textbf{62.4} & \textit{67.5} & \textit{72.8} & \textit{71.5} & \textit{63.9} & \textit{49.0} & \textit{65.8} \\

\hline


%    \hline
    \multicolumn{1}{c|}{\multirow{15}[2]{*}{\textbf{MT(62)}}} & \multicolumn{1}{c|}{\multirow{5}[2]{*}{NM\#5-6}} & AE    & 49.3  & 61.5  & 64.4  & 63.6  & 63.7  & 58.1  & 59.9  & 66.5  & 64.8  & 56.9  & 44.0  & 59.3  \\
%\cline{3-15}
          &       & MGAN  & 54.9  & 65.9  & 72.1  & 74.8  & 71.1  & 65.7  & 70.0  & 75.6  & 76.2  & 68.6  & 53.8  & 68.1  \\
%\cline{3-15}
          &       & GaitSet & 89.7  & 97.9  & 98.3  & 97.4  & 92.5  & 90.4  & 93.4  & 97.0  & 98.9  & 95.9  & 86.6  & 94.3  \\
%\cline{3-15}
&       & GaitGL (baseline)   & \textit{93.9} & \textit{97.6} & \textit{98.8} & \textit{97.3} & \textit{95.2} & \textit{92.7} & \textit{95.6} & \textit{98.1} & \textit{98.5} & \textit{96.5} & \textit{91.2} & \textit{95.9} \\
&       & \textbf{Ours} (${\mathcal{L}_m}$)  & 90.7 & 96.6 & 98.2 & 96.7 & 93.0 & 91.4 & 94.0 & 97.1 & 97.1 & 94.9 & 86.9 & 94.2 \\
&       & \textbf{Ours} (${\mathcal{L}_s}$)  & \textbf{94.2} & \textbf{98.3} & \textbf{99.0} & \textbf{97.5} & \textbf{95.7} & \textbf{94.4} & \textbf{96.8} & \textbf{98.7} & \textbf{99.0} & \textbf{97.4} & \textbf{92.5} & \textbf{96.7} \\

\cline{2-15}          & \multicolumn{1}{c|}{\multirow{5}[2]{*}{BG\#1-2}} & AE    & 29.8  & 37.7  & 39.2  & 40.5  & 43.8  & 37.5  & 43.0  & 42.7  & 36.3  & 30.6  & 28.5  & 37.2  \\
%\cline{3-15}
          &       & MGAN  & 48.5  & 58.5  & 59.7  & 58.0  & 53.7  & 49.8  & 54.0  & 51.3  & 59.5  & 55.9  & 43.1  & 54.7  \\
%\cline{3-15}
         &       & GaitSet & 79.9  & 89.8  & 91.2  & 86.7  & 81.6  & 76.7  & 81.0  & 88.2  & 90.3  & 88.5  & 73.0  & 84.3  \\
%\cline{3-15}
% &       & Ours  & \RED{85.6} & \RED{93.2} & \RED{95.0} & \RED{92.4} & \RED{89.0} & \RED{81.5} & \RED{86.6} & \RED{92.7} & \RED{95.6} & \RED{92.9} & \RED{83.1} & \RED{89.8} \\
&       & GaitGL (baseline)   & 88.5 & 95.1 & 95.9 & 94.2 & \textit{91.5} & 85.4 & 89.0 & 95.4 & \textit{97.4} & \textit{94.3} & \textit{86.3} & 92.1 \\
&       & \textbf{Ours} (${\mathcal{L}_m}$)  & \textit{89.2} & \textit{96.2} & \textit{97.0} & \textit{94.4} & 91.3 & \textit{87.5} & \textit{90.7} & \textit{95.4} & 96.4 & 93.2 & 81.9 & \textit{92.1} \\
&       & \textbf{Ours} (${\mathcal{L}_s}$)  & \textbf{90.9} & \textbf{95.7} & \textbf{97.3} & \textbf{96.5} & \textbf{92.4} & \textbf{89.4} & \textbf{92.3} & \textbf{96.4} & \textbf{98.0} & \textbf{95.8} & \textbf{89.0} & \textbf{94.0} \\

\cline{2-15}          & \multicolumn{1}{c|}{\multirow{5}[2]{*}{CL\#1-2}} & AE    & 18.7  & 21.0  & 25.0  & 25.1  & 25.0  & 26.3  & 28.7  & 30.0  & 23.6  & 23.4  & 19.0  & 24.2  \\
%\cline{3-15}
          &       & MGAN  & 23.1  & 34.5  & 36.3  & 33.3  & 32.9  & 32.7  & 34.2  & 37.6  & 33.7  & 26.7  & 21.0  & 31.5  \\
%\cline{3-15}
         &       & GaitSet & 52.0  & 66.0  & 72.8  & 69.3  & 63.1  & 61.2  & 63.5  & 66.5  & 67.5  & 60.0  & 45.9  & 62.5  \\
%\cline{3-15}
% &       & Ours  & \RED{70.2} & \RED{83.6} & \RED{87.3} & \RED{85.2} & \RED{78.5} & \RED{73.1} & \RED{80.0} & \RED{85.1} & \RED{84.6} & \textbf{76.9} & \textbf{61.7} & \textbf{78.7} \\
&       & GaitGL (baseline)   & 70.7 & 83.2 & 87.1 & 84.7 & 78.2 & 71.3 & {78.0} & {83.7} & 83.6 & 77.1 & 63.1 & 78.3 \\
&       & \textbf{Ours} (${\mathcal{L}_m}$)  & \textbf{76.3} & \textbf{91.5} & \textbf{94.8} & \textbf{90.7} & \textbf{86.4} & \textbf{80.8} & \textbf{84.8} & \textbf{90.7} & \textbf{91.0} & \textit{82.8} & \textit{65.3}  & \textbf{85.0} \\
&       & \textbf{Ours} (${\mathcal{L}_s}$)  & \textit{73.6} & \textit{88.0} & \textit{92.7} & \textit{88.3} & \textit{82.4} & \textit{76.8} & \textit{82.6} & \textit{88.8} & \textit{89.3} & \textbf{83.7} & \textbf{67.3} & \textit{83.0} \\

    \hline

%    \hline
    \multicolumn{1}{c|}{\multirow{17}[2]{*}{\textbf{LT(74)}}} & \multicolumn{1}{c|}{\multirow{7}[2]{*}{NM\#5-6}} & CNN-3D & 87.1  & 93.2  & 97.0  & 94.6  & 90.2  & 88.3  & 91.1  & 93.8  & 96.5  & 96.0  & 85.7  & 92.1  \\
%\cline{3-15}
          &       & CNN-Ensemble & 88.7  & 95.1  & 98.2  & 96.4  & 94.1  & 91.5  & 93.9  & 97.5  & 98.4  & 95.8  & 85.6  & 94.1  \\
%\cline{3-15}
          &       & GaitSet & 91.1  & 99.0 & 99.9 & 97.8  & 95.1  & 94.5  & 96.1  & 98.3  & 99.2 & 98.1  & 88.0  & 96.1  \\
%\cline{3-15}
&       & ACL   & 92.0  & 98.5  & \textbf{100.0} & \textbf{98.9} & 95.7  & 91.5  & 94.5  & 97.7  & 98.4  & 96.7  & 91.9  & 96.0  \\
&       & GaitPart & 94.1  & 98.6 & 99.3  & 98.5  & 94.0  & 92.3  & 95.9  & 98.4  & 99.2 & 97.8  & 90.4  & 96.2  \\
&       & GaitGL (baseline)  & \textit{96.0} & \textit{98.3}  & \textit{99.0}  & 97.9  & \textbf{96.9} & 95.4 & \textit{97.0} & \textit{98.9} & \textit{99.3} & \textbf{98.8} & \textbf{94.0} & \textit{97.4} \\
&       & \textbf{Ours} (${\mathcal{L}_m}$)  & 95.2 & 97.4 & 98.4 & 97.3 & 95.4 & 94.4 & 95.9 & 98.5 & 98.4 & 97.9 & 91.9 & 96.4 \\
&       & \textbf{Ours} (${\mathcal{L}_s}$)  & \textbf{96.8} & \textbf{98.8} & 99.2 & \textit{98.3} & \textit{96.3} & \textbf{96.2} & \textbf{97.8} & \textbf{99.0} & \textbf{99.4} & \textit{98.7} & \textit{93.2} & \textbf{97.6} \\

\cline{2-15}
          & \multicolumn{1}{c|}{\multirow{5}[2]{*}{BG\#1-2}} & CNN-LB & 64.2  & 80.6  & 82.7  & 76.9  & 64.8  & 63.1  & 68.0  & 76.9  & 82.2  & 75.4  & 61.3  & 72.4  \\
%\cline{3-15}
          &       & GaitSet & 86.7  & 94.2  & 95.7  & 93.4  & 88.9  & 85.5  & 89.0  & 91.7  & 94.5  & 95.9  & 83.3  & 90.8  \\
%\cline{3-15}
&       & GaitPart & 89.1  & 94.8 & 96.7 & 95.1 & 88.3  & 84.9 & 89.0  & 93.5  & 96.1  & 93.8  & 85.8  & 91.5  \\
&       & GaitGL (baseline)  & \textit{92.6} & 96.6 & 96.8 & 95.5 & 93.5 & 89.3 & 92.2 & 96.5 & \textit{98.2} & \textit{96.9} & \textit{91.5} & 94.5 \\
&       & \textbf{Ours} (${\mathcal{L}_m}$)  & 91.5 & \textit{96.9} & \textbf{97.8} & \textit{96.3} & \textit{94.5} & \textbf{91.8} & \textbf{93.9} & \textit{96.7} & 98.0 & \textbf{97.4} & 90.4 & \textit{95.0} \\
&       & \textbf{Ours} (${\mathcal{L}_s}$)  & \textbf{93.0} & \textbf{97.5}  & \textit{97.4} & \textbf{97.3} & \textbf{95.8} & \textit{91.7} & \textit{93.5} & \textbf{97.7} & \textbf{98.2} & 96.6 & \textbf{91.7} & \textbf{95.5} \\

\cline{2-15} & \multicolumn{1}{c|}{\multirow{5}[2]{*}{CL\#1-2}} & CNN-LB & 37.7  & 57.2  & 66.6  & 61.1  & 55.2  & 54.6  & 55.2  & 59.1  & 58.9  & 48.8  & 39.4  & 54.0  \\
%\cline{3-15}
          &       & GaitSet & 59.5  & 75.0  & 78.3  & 74.6  & 71.4  & 71.3  & 70.8  & 74.1  & 74.6  & 69.4  & 54.1  & 70.3  \\
% \cline{3-15}
&       & GaitPart & 70.7  & 85.5  & 86.9  & 83.3  & 77.1  & 72.5  & 76.9  & 82.2  & 83.8  & 80.2  & 66.5  & 78.7  \\
&       & GaitGL (baseline)  & 76.6 & 90.0 & 90.3 & 87.1 & 84.5 & 79.0 & 84.1 & 87.0 & 87.3 & 84.4 & 69.5 & 83.6 \\
&       & \textbf{Ours}(${\mathcal{L}_m}$) & \textbf{82.9} & \textbf{95.2} & \textbf{97.3} & \textbf{94.0} & \textbf{91.2} & \textbf{84.8} & \textbf{88.2} & \textbf{92.9} & \textbf{92.9} & \textbf{89.9} & \textbf{78.1} & \textbf{89.8} \\
&       & \textbf{Ours} (${\mathcal{L}_s}$)  & \textit{76.8} & \textit{92.8} & \textit{94.4} & \textit{91.5} & \textit{87.6} & \textit{82.3} & \textit{87.7} & \textit{91.2} & \textit{92.5} & \textit{87.8} & \textit{74.1} & \textit{87.2} \\
    \bottomrule
    \end{tabular}%
    }
%   \label{tab:addlabel}%
\label{tab:sota_casiab}
\vspace*{-2em}
\end{table*}%
\end{center}

\vspace{-2em}
\section{Experiments}

\begin{table*}[htbp]
  \centering
  \vspace{-1em}
  \caption{Rank-1 accuracy (\%) on OUMVLP under 14 probe views excluding identical-view cases.}
  \vspace{-1em}
  \resizebox{0.98\textwidth}{!}{
    \begin{tabular}{c|c|c|c|c|c|c|c|c|c|c|c|c|c|c|c}
    \toprule
    \multirow{2}[2]{*}{\textbf{Method}} & \multicolumn{14}{c|}{\textbf{Probe View}}                                                            & \multicolumn{1}{c}{\multirow{2}[2]{*}{\textbf{Mean}}} \\
\cline{2-15}    \multicolumn{1}{c|}{} & $0^{\circ}$  & $15^{\circ}$  & $30^{\circ}$  & $45^{\circ}$  & $60^{\circ}$  & $75^{\circ}$  & $90^{\circ}$ & $180^{\circ}$  & $195^{\circ}$  & $210^{\circ}$  & $225^{\circ}$  & $240^{\circ}$  & $255^{\circ}$  & $270^{\circ}$  &  \\
    \midrule
    GEINet & 23.2  & 38.1  & 48.0  & 51.8  & 47.5  & 48.1  & 43.8  & 27.3  & 37.9  & 46.8  & 49.9  & 45.9  & 45.7  & 41.0  & 42.5  \\
    \hline
    GaitSet & 79.3  & 87.9  & 90.0  & 90.1  & 88.0  & 88.7  & 87.7  & 81.8  & 86.5  & 89.0  & 89.2  & 87.2  & 87.6  & 86.2  & 87.1  \\
    \hline
    GaitPart & 82.6  & 88.9  & 90.8  & 91.0  & 89.7  & 89.9  & 89.5  & 85.2  & 88.1  & 90.0  & 90.1  & 89.0  & 89.1  & 88.2  & 88.7  \\
    \hline
    GLN   & 83.8  & 90.0  & 91.0  & 91.2  & 90.3  & 90.0  & 89.4  & 85.3  & 89.1 & 90.5 & 90.6 & 89.6  & 89.3  & 88.5  & 89.2  \\
    \hline
    GaitGL(baseline)  & 84.9 & 90.2 & 91.1 & 91.5 & 91.1 & 90.8 & 90.3 & 88.5 & 88.6  & 90.3  & 90.4  & 89.6 & 89.5 & 88.8 & 89.7 \\
    \hline 
    \textbf{Ours}  & \textbf{87.1} & \textbf{91.0} & \textbf{91.4} & \textbf{91.7} & \textbf{91.5} & \textbf{91.3} & \textbf{91.0} & \textbf{90.1} & \textbf{89.8}  & \textbf{90.6}  & \textbf{90.7} & \textbf{90.3} & \textbf{90.2} & \textbf{89.8} & \textbf{90.5} \\
    \bottomrule
    \end{tabular}%
    }
    %\vspace*{-1em}
  \label{comparision_oumvlp}%
\end{table*}%

% \begin{table*}[t]
% 	\small
% 	\begin{center}
% 		\centering
% 		\setlength{\tabcolsep}{8pt}
% 		\caption{
% 			The hyperparameters used in our experiments. BS, Lr init, Lr min, Scheduler are the batch size, initial learning rate, final learning rate and learning rate scheduler. OPT, Momentum and W decay are optimizer, the optmizer momentum and the weight decay. 
% 			Warmup Iters, Warmup Lr and Iters means the warmup iteration, warmup learning rate and total iteration.
% 		}
% 		\label{tab:paras}
% 			\begin{tabular}{ c | c | c | c | c | c | c | c | c | c | c } 
% 				\toprule[1.2pt]
% 				Dataset  & BS  & Lr init & Lr min & Scheduler & OPT & Momentum & W decay & Warmup Iters  & Warmup lr & Iters    \\
% 				\hline
% 				CASIA-B  & [8, 8] & 0.1 & 1e-4 & Cosine   & SGD & 0.9 & 5e-4 & 1k & 1e-6 & 80k  \\
% 				OUMVLP & [32, 8] & 0.1  & 1e-4 & Cosine & SGD & 0.9 & 5e-4 & 2k & 1e-6  & 210k   \\
% 				\bottomrule[1.2pt]
% 			\end{tabular}
		
% 	\end{center}

% \end{table*}

\subsection{Datasets}
We conduct our experiments on two commonly used public datasets CASIA-B and OUMVLP.
\begin{figure}[t]
	\centering
	%\vspace{0.3cm}
	\includegraphics[width=0.5\textwidth,height=2.5cm]{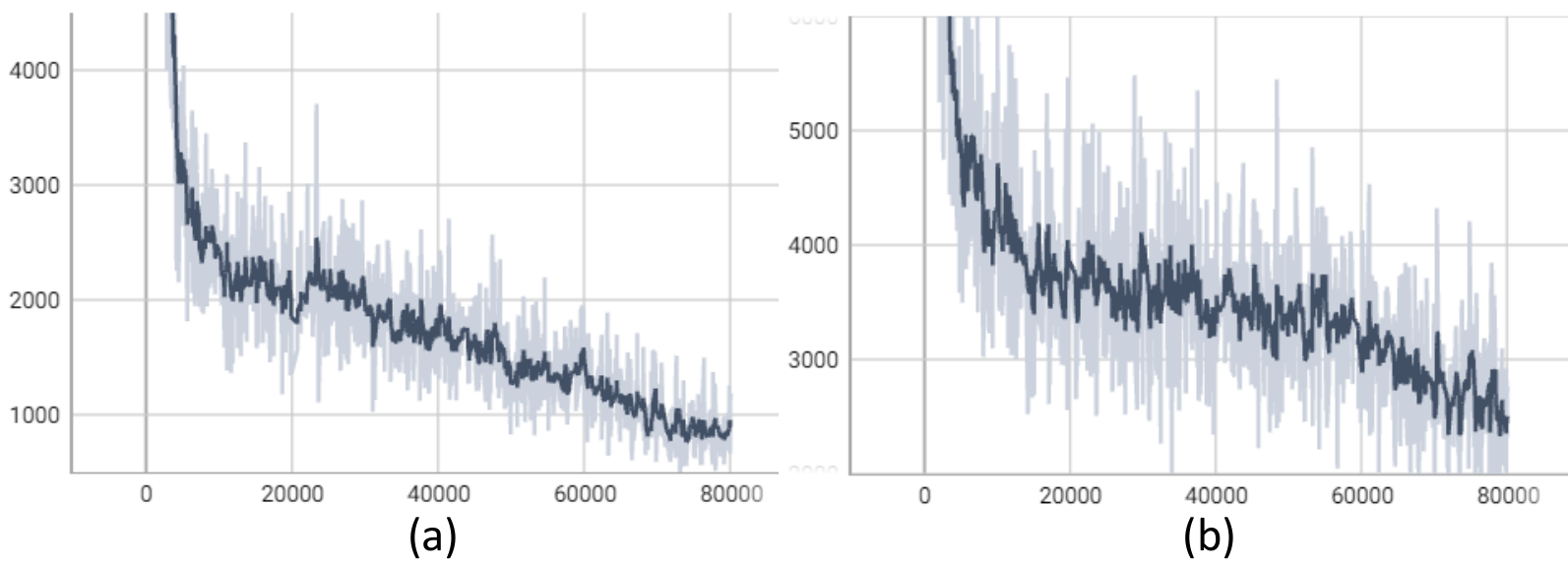}
	\caption{Loss tendency from iteration 0 to iteration 80k. Y-axis indicates the none-zero numbers of triplets in a batch. (a) baseline. (b) the proposed method.}
	\label{fig:loss}
	\vspace{-2em} 
\end{figure}
\\\textbf{CASIA-B.}  The CASIA-B dataset \cite{6115889} is the popular cross-cloth gait database. It includes 124 subjects, each of which has 10 groups of videos. Among these groups, six of them are sampled in normal walking (NM), two groups are in walking with a bag (BG), and the rest are in walking in different cloth (CL). Each group contains 11 gait sequences from different angles ($0^{\circ}$-$180^{\circ}$
and the sampling interval is $18^{\circ}$). Therefore, there are 124 (subject) × 10 (groups) × 11 (view
angle) = 13,640 gait sequences in CASIA-B. The gait sequences of each subject are divided into train set and
test set. Following the setting of previous works \cite{chao2021gaitset}, in small-sample training (ST) the first
24 subjects (labeled in 001-024) are used for training and the
rest 100 subjects are left for testing. In medium-sample
training (MT), the first 62 subjects are used for training and the rest 62 subjects are left for test. In large-sample training (LT), the first 74 subjects are used for training and
the rest 50 subjects are leaved for test. In the test stage,
the sequences NM\#01-NM\#04 are taken as the gallery set,
while the sequences NM\#05-NM\#06, BG\#01-BG\#02, and
CL\#01-CL\#02 are considered as the probe set to evaluate
the performance.
\\\textbf{OUMVLP.} The OUMVLP \cite{takemura2018multi} dataset is one of the largest gait recognition open source databases, which contains 10,307 subjects in total. Each subject contains two groups of videos, Seq\#00 and Seq\#01. Each group of sequences is captured from 14 angles ($0^{\circ}$-$90^{\circ}$, $180^{\circ}$-$270^{\circ}$ and the sampling interval is $15^{\circ}$).
Following the setting of previous works \cite{chao2021gaitset}, We adopt the same protocol, i.e., 5,153 subjects are taken as training data and 5,154 subjects are used as test data to evaluate the performance of the proposed method. In the test stage, the sequences in Seq\#01 are taken as the gallery set, while the sequences in Seq\#00 are regarded as the probe set for evaluation.
%\\\textbf{Evaluation metrics}
%In the testing phase, we compare the feature similarities between probe and gallery samples to identify a person and report the performance of the average Rank-1 recognition accuracy.

% \begin{figure}[t]
% 	\centering
% 	%\vspace{0.3cm}
% 	\includegraphics[width=0.5\textwidth]{fig/loss2.pdf}
% 	\caption{Loss tendency from iteration 0 to iteration 80000. The first row is baseline, the second row indicates}
% 	\label{fig:loss}
% 	% \vspace{-1em} 
% \end{figure}

\begin{figure}[t]
	\centering
	% \vspace{-0.3cm}
	\includegraphics[width=0.5\textwidth,height=3.5cm]{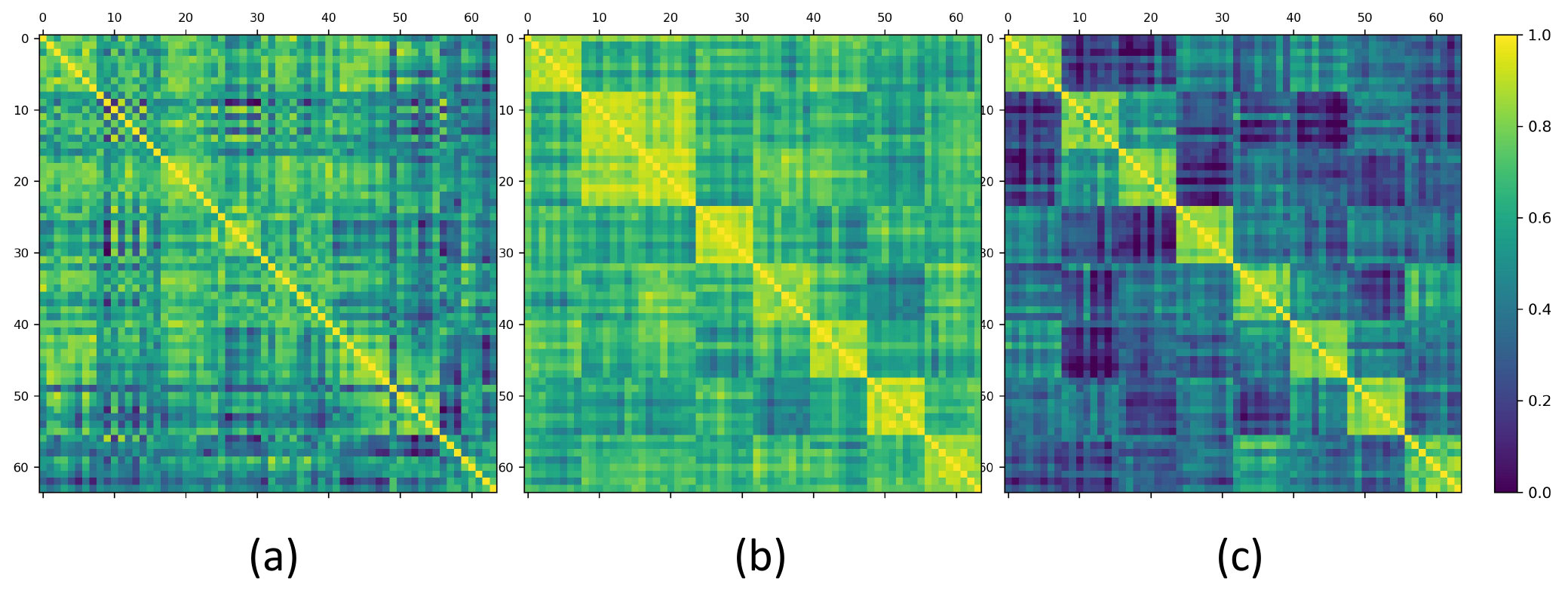}
	\vspace{-1em}
	\caption{Visualization of Similarity Matrix. The matrix size is 64*64, and batch size is [8,8]. (a) similarity matrix at iteration 1k. (b) similarity matrix at iteration 40k. (c) similarity matrix at iteration 80k. }
	\label{fig:vis_sim}
	\vspace{-1em} 
\end{figure}

\subsection{Implementation Details}
The baseline network is GaitGL \cite{lin2021gait}, which utilizes global and local convolutional layers to extract global and local features, and uses local temporal aggregation to process the temporal information. We adopt the same preprocessing approach as \cite{chao2019gaitset} to obtain gait silhouettes for CASIA-B and OUMVLP. The loss function coefficients are all set to 1 in our experiments. We adopt the SGD optimizer \cite{SGD} with 0.1 learning rate. The momentum and weight decay are set to 0.9 and 5e-4, respectively. The learning rate decays from 0.1 to 1e-4 with a cosine scheduler. We implement our method on the basis of OpenGait\footnote{\href{https://github.com/ShiqiYu/OpenGait.git}{https://github.com/ShiqiYu/OpenGait.git}}. 
\\\textbf{CASIA-B} In the setting of CASIA-B experiments, the iteration number is set to 80K. In each batch, the number of subjects and the number of sequences for each subject are set to (8, 8) and the input resolution is (64, 44). For evaluation, all silhouettes of gait sequences are used to obtain the final representation. In the CASIA-B ST, MT and LT settings, the iteration number is set to 60K, 80K and 80K, respectively.
\\\textbf{OUMVLP} In the setting of OUMVLP experiments, the iteration number is set to 210K. The number of subjects and the number of sequences for each subject are set to (32, 8) and the input resolution is (64, 44). For evaluation, all silhouettes of gait sequences are taken to obtain the final representation. 
\begin{table*}[!t]
	\small
	\centering
	\setlength{\tabcolsep}{10pt}
	\vspace{-1em}
	\caption{Component-wise analysis. Ablation study on CASIA-B, excluding identical-view cases.}
%    \begin{center}
    \vspace{-1em}
		\begin{tabular}{c|cccc|cccc}
		
			% \toprule[1.07pt]
			\hline
			 & Data  & s-triplet & SimCE   & m-SimCE &  \multicolumn{4}{c}{CASIA-B Dataset} \\
			 &  Augmentation  & Loss  & Loss  & Loss  & NM $\bm{\uparrow}$ & BG $\bm{\uparrow}$    & CL $\bm{\uparrow}$  & Mean $\bm{\uparrow}$\\
			% \midrule[1.07pt]
			% \hline
			 \hline
			% \textbf{Variants} & \textbf{1 shot} & \textbf{5 shot} \\
			 \#1 & & & & & 97.4  & 94.6 & 83.8 & 91.9 \\
			 \#2 &\cmark  & & & & 97.2  & 95.2 & 85.6 & 92.6 \\
			 \#3 &\cmark     &  \cmark & &  &  97.8  & 95.8 & 87.0 & 93.5  \\
%			 \#4 &\cmark     &   & \cmark &  &  96.9  & 93.3 & 80.6 & 90.3 \\
%			 \#5 &\cmark     &   &  & \cmark &  94.7  & 91.9 & 82.8 &  89.9 \\
			 \#4 &\cmark &  \cmark & & \cmark &  96.4  & 95.0 & 89.8 & 93.7 \\
			 \#5 &\cmark   &  \cmark &  \cmark &  &  97.2  & 95.7 & 88.5 & 93.8\\
			\hline
			% \bottomrule[1.07pt]
		\end{tabular}
%	\end{center}
	\label{tab:ablation}
	\normalsize
	%\vspace{-1em}
\end{table*}
\begin{table}[t]  
	\centering
	\small
	\setlength{\tabcolsep}{9pt}
	\vspace{-1em}
	% \fontsize{10}{9}\selectfont
	\caption{Model agnostic results on CASIA-B dataset, excluding identical-view cases.}
	\label{tab:agnostic}
	\vspace{-1em}
			\begin{tabular}{lccccc}
				\toprule[1.1pt]
				Methods  & NM $\bm{\uparrow}$ & BG  $\bm{\uparrow}$ & CL    $\bm{\uparrow}$ & Mean    $\bm{\uparrow}$ 
				\\ \midrule[1.1pt]
				OpenGait &$96.3$ & $92.2$ & $77.6$ & $88.7$  \\
				\textbf{+Ours} (${\mathcal{L}_s}$) & \textbf{97.9} & \textbf{93.2} & \textbf{78.7} & \textbf{90.0}  \\
				%\textbf{+Ours} (Contrastive)  & 97.4 & \textbf{94.0} & \textbf{79.4}  \\
				\midrule
				GaitSet~\cite{chao2021gaitset} &$95.8$ & $90.0$ & $75.4$ & $87.0$ \\
				\textbf{+Ours} (${\mathcal{L}_s}$) & \textbf{97.0} & \textbf{92.6} & \textbf{76.6} & \textbf{88.7}  \\
				\midrule
				GaitGL~\cite{lin2021gait} & $97.4$ & $94.5$ & $83.6$ & $91.8$  \\			
				\textbf{+Ours} (${\mathcal{L}_s}$) & \textbf{97.6} & \textbf{95.5} & \textbf{87.2} & \textbf{93.4} \\
				% \textbf{+Ours} (Contrastive)  & 96.4 & 95.0 & \textbf{89.8} \\
				\bottomrule[1.1pt]
		\end{tabular} \label{tab:Model agnostic}
		\vspace{-2em}
\end{table}
\subsection{Results}
\subsubsection{Comparison with State-of-the-art Methods}
\textbf{CASIA-B.} We compare our approach with state-of-the-art gait recognition approaches, including ViDP \cite{hu2013view}, CMCC \cite{kusakunniran2013recognizing}, GaitSet \cite{chao2019gaitset}, AE \cite{yu2017invariant}, MGAN \cite{he2018multi}, CNN-LB, CNN-3D, CNN-Ensemble \cite{wu2016comprehensive}, ACL \cite{zhang2019cross}, GaitPart \cite{fan2020gaitpart} and GaitGL \cite{lin2021gait} on CASIA-B. The experimental results are shown in Table.\ref{tab:sota_casiab}, and we summarize the results as follows: 1) the proposed approach achieves the best mean recognition accuracy and at nearly all view metrics. \textbf{Both ${\mathcal{L}_s}$ and ${\mathcal{L}_m}$ improve the baseline by a large margin.} The experiment results demonstrate the effectiveness of the proposed method. 2) we find that ${\mathcal{L}_s}$ performs better on NM and BG conditions while ${\mathcal{L}_m}$ surpasses other methods with a great performance margin in CL scenario. The reason can refer to Sec.\ref{sec:optimization range}, the rise in CL is due to the advantages of generalized ${\mathcal{L}_m}$, where it considers more terms to find a better optimization direction, while its inferior in NM is mainly due to the disadvantages in Sec.\ref{sec:optimization range}. And our guideline is that if the CL makes up a high proportion in the dataset, we recommend using ${\mathcal{L}_m}$, and otherwise using ${\mathcal{L}_s}$. 3) in LT setting CL condition, the proposed method on CL condition reaches 89.8\%, which is a significant performance, and the gap between NM and CL condition decreases from 13.8\% to 6.6\%, which indicates that although large appearance changes bring big data variations, CL variance can be alleviated by loss design. 4) as illustrated in Fig.\ref{fig:vis_sim}, the similarity matrix between a batch $[8,8]$ in 1k iteration, 40k iteration and 80k iteration, the 8 diagonal $8*8$ blocks indicates the intra-class similarity, and the 56 off-diagonal blocks indicate the inter-class similarity. The similarity matrix at iteration 1k is close to randomly distributed features, and after 40k iteration training, the similarity intra-class increases and the diagonal blocks get closer to 1, and after more iterations training, the inter-class similarity decreases so that the off-diagonal blocks get closer to 0, indicating that the feature representations are discriminative. 5) as shown in Fig.\ref{fig:loss}(b), the triplet loss number $N_{non}$ of proposed generalized similarity loss is larger than baseline, which indicates the overfitting is alleviated.  Therefore, the proposed method exploits a better metric in gait recognition compared with SOTA methods.
\\\textbf{OUMVLP} As shown in Tab.\ref{comparision_oumvlp}, we also compare our approach with state-of-the-art gait recognition approaches on OUMVLP, including GaitNet \cite{song2019gaitnet}, GaitSet \cite{chao2019gaitset}, GaitPart \cite{fan2020gaitpart}, GLN \cite{hou2020gait} and GaitGL \cite{lin2021gait}. Note that we adopt GaitGL as our baseline and adopt ${\mathcal{L}_s}$ according to the guidelines, and the results outperform state-of-the-art gait by a large margin. The results again demonstrate the effectiveness of the proposed metric.
\subsubsection{Model Agnostic Results} As shown in Tab.\ref{tab:agnostic}, we conduct experiments on OpenGait baseline, GaitSet \cite{chao2019gaitset}, GaitGL \cite{lin2021gait}, and the results show that the proposed metric can improve the performances of the baseline at all metrics consistently, regardless of baseline backbones or network structures. The model agnostic property further verifies that the proposed loss outperforms other popular losses in gait recognition.

% \begin{table*}[!t]
% 	\small
% 	\centering
% 	\setlength{\tabcolsep}{10pt}
% 	\caption{\small{Component-wise analysis. Ablation study on CASIA-B.}}
% %    \begin{center}
% 		\begin{tabular}{c|ccccc|ccc}
		
% 			% \toprule[1.07pt]
% 			\hline
% 			 & RandomErase & Dilate & AutoHM & SimCE  & Contrastive &  \multicolumn{3}{c}{CASIA-B Dataset} \\
% 			 &  Augmentation &  Augmentation & Module  & Loss  & Loss  & NM $\bm{\uparrow}$ & BG $\bm{\uparrow}$    & CL $\bm{\uparrow}$ \\
% 			% \midrule[1.07pt]
% 			% \hline
% 			 \hline
% 			% \textbf{Variants} & \textbf{1 shot} & \textbf{5 shot} \\
% 			 \#1 & & & & & & 97.4  & 94.6 & 83.8 \\
% 			 \#2 & \cmark & &  & & & 97.6  & 95.3 & 84.2 \\
% 			 \#3 &  & \cmark &  & & & 96.9  & 94.5 & 84.6 \\

% 			 \#4 &\cmark  &   &  \cmark & &  &  \textbf{98.1}  & \textbf{95.8} & 85.6 \\
% 			 \#5 &\cmark  &   &   & \cmark &  &  96.9  & 93.3 & 80.6 \\
% 			 \#6 &\cmark  &   &   &  & \cmark &  94.7  & 91.9 & 82.8 \\
% 			 \#7 &\cmark  & \cmark  &  \cmark & &  &  97.8  & \textbf{95.8} & 87.0 \\
% 			 \#8 &\cmark  &   &  \cmark &  \cmark &  &  97.6  & 95.5 & 87.2 \\
% 			 \#9 &\cmark  &  \cmark  &  \cmark &  \cmark &  &  97.2  & 95.7 & 88.5 \\

% 			 \hline
% 			 %\hline
% 			 % \multicolumn{1}{c|}{\multirow{1.5}[2]{*}{Self-supervised}} & & \cmark & &  & 90.3  & 81.0 & 37.0 \\
% 			 \#10 &\cmark  &  \cmark &  \cmark & & \cmark &  96.4  & 95.0 & \textbf{89.8} \\
% 			\hline
% 			% \bottomrule[1.07pt]
% 		\end{tabular}
% %	\end{center}
% 	\label{tab:ablation}
% 	\normalsize
% \end{table*}
\subsubsection{Ablation Study} The ablation result is illustrated in Tab.\ref{tab:ablation}. Note that the data augmentation is firstly sampled from many data augmentation methods, and then is fixed (in experiments they are random erase and image dilate) for all experiments from \#2 to \#5 for a fair comparison. Here is the analysis. 1) from experiment \#2, data augment improves baseline by less than 1\%, which is reasonable. 2) compare \#3 with \#2, replacing triplet loss by the proposed s-triplet loss (Eq.\ref{eq:s-triloss}) brings a consistent rise at all metrics. 3) experiments \#4 is the proposed ${\mathcal{L}_s}$, which improves the CL condition, while the NM and BG stay nearly the same. The experiment \#5 is the proposed ${\mathcal{L}_m}$, which improves CL by a large margin, by NM and BG declines, and the reason for this is illustrated in Sec.\ref{sec:optimization range}. (4) Overall, the mean accuracy rises by about 2\%, which demonstrates the effectiveness of the proposed methods.

\section{Conclusion}
In this work, we propose a generalized similarity loss to resolve the small inter-class variance problem in gait recognition. We analyze the properties of the proposed method from three aspects, namely inter-class hard mining, uniformity and robustness of inter-class feature distribution, and dynamic margin, and how these aspects help with inter-class feature distribution. Extensive experiments on CASIA-B and OUMVLP demonstrate that the proposed loss achieves state-of-the-art performance greatly regardless of network architecture in gait recognition.

% \section{Acknowledgments}

% \section{Appendices}


\bibliographystyle{ACM-Reference-Format}
\bibliography{ref}

\appendix

\end{document}